\documentclass{IEEEmce}

\usepackage[colorlinks,urlcolor=blue,linkcolor=blue,citecolor=blue]{hyperref}
\usepackage{cite}
\usepackage{graphicx}
\usepackage{subfigure}
\usepackage{url}
\usepackage{siunitx}
\usepackage{gensymb}
\usepackage{textcomp}
\usepackage[numbers,sort&compress]{natbib}
\usepackage{amsmath}
\usepackage{upmath}

\jvol{XX}
\jnum{XX}
\paper{8}
\jmonth{May/June}
\publisheddate{00 xxxx 0000}
\currentdate{00 xxxx 0000}
\jname{IEEE Consumer Electronics Magazine}
\pubyear{2024}
\doiinfo{MCE.2024.Doi Number}

\begin{document}

\sptitle{Department: AI Home Robots}
\editor{Editor: Name, xxxx@email}

\title{Bringing Robots Home: The Rise of AI Robots in Consumer Electronics}

\author{Haiwei Dong}
\affil{Huawei Canada, University of Ottawa}

\author{Yang Liu}
\affil{Shopify Inc., University of Ottawa}

\author{Ted Chu}
\affil{Millennium Institute}

\author{Abdulmotaleb El Saddik}
\affil{University of Ottawa, MBZUAI}

\markboth{Department Head}{Paper title}

\maketitle

\enlargethispage{10pt}

\begin{figure*}[]
\centering
\includegraphics[width=\textwidth, trim={0 0 0 0},clip]{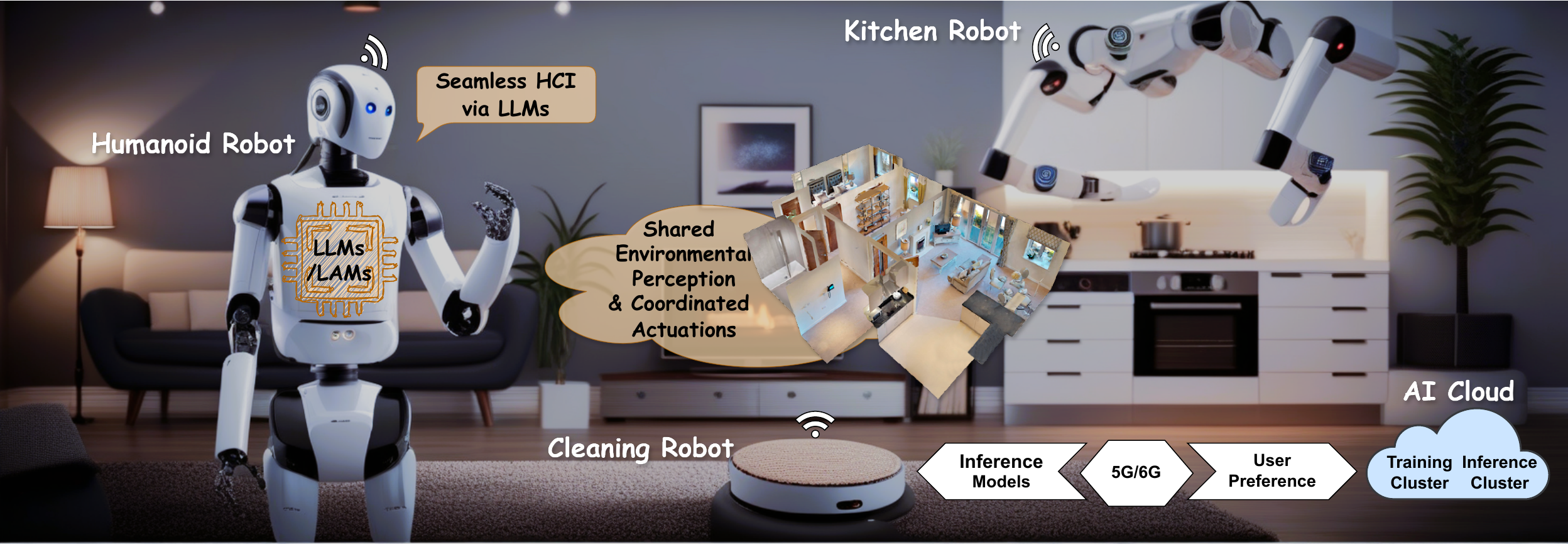}
   \caption{Effortless information exchange among a humanoid robot, a kitchen robot, and a cleaning robot is facilitated, with all robots accessing the latest inference models from the AI cloud based on user preferences.}
\label{fig:framework}
\end{figure*}



\chapterinitial{O}n March 18, 2024, NVIDIA unveiled Project GR00T, a general-purpose multimodal generative AI model designed specifically for training humanoid robots \cite{nvidia_GR00T}. Preceding this event, Tesla's revealing of the Optimus Gen 2 humanoid robot on December 12, 2023, underscored the profound impact robotics is poised to have on reshaping various facets of our daily lives \cite{openai_optimus}. While robots have long dominated industrial settings, their presence within our homes is a burgeoning phenomenon. This can be attributed, in part, to the complexities of domestic environments and the challenges of creating robots that can seamlessly integrate into our daily routines.

However, significant advancements in Artificial Intelligence (AI) are bridging this gap. AI empowers robots to navigate dynamic environments, understand user commands, and even learn and adapt over time. This confluence of AI and robotics is ushering in a new era of intelligent home robots. We are witnessing a surge in affordable, user-friendly robots specifically designed to tackle everyday tasks. Robotic vacuums tirelessly maintain our floors, while lawnmowers manage our yards with unwavering precision. Even rudimentary kitchen assistants are emerging, assisting with tasks like chopping vegetables, mixing ingredients, and streamlining meal prep \cite{moley_kichen}.

The impact of AI extends beyond these task-specific robots. The trend towards human-like robots with advanced AI capabilities is transforming how we envision in-home assistance. Driven by factors like the global pandemic and population aging, there's a growing need for robots that can provide targeted support, such as nursing robots \cite{riken_riba}. These AI-powered robots can offer remote health monitoring, medication reminders, and even basic physical assistance, empowering elderly to maintain independence and enhance their quality of life. Additionally, these robots can provide companionship and alleviate social isolation, particularly for those living alone or in remote areas.

\section{Constituents of a Robotic Body}

Just like any living organism, home robots rely on a complex network of sensors and actuators to function effectively within our dynamic domestic environments. These components work in tandem to gather information about the surroundings and translate that information into physical actions.

\textbf{Sensory Perception: The Senses of a Robot.} Sensors act as the robot's eyes, ears and skin, constantly gathering data about its surroundings. Here's a breakdown of some key sensor types in home robots:

\begin{itemize}
\item Vision Sensors: Multimodal cameras empower robots with heightened perception, facilitating precise tasks such as object recognition and navigation within their surroundings. Advanced robots might even utilize specialized navigation sensors  (such as RGB-D cameras, event cameras, infrared cameras, LiDAR or ultrasonic arrays) to help robots map their surroundings and navigate autonomously.
\item Audio Sensors: Microphones enable robots to detect and interpret sounds. This can be crucial for responding to voice commands, identifying potential hazards (like alarms), and even gauging user emotions based on vocal cues.
\item Movement Sensors: Gyroscopes and accelerometers track the robot's position and movement, allowing for precise control and stable operation.
\item Touch/Force Sensors: These physical interaction sensors use piezoelectricity to measure electrical charges due to mechanical deformations (pressure, bending) along a specific direction. The haptic feedback/textural sensations originate from them.
\end{itemize}

\textbf{Taking Action: The Muscles of a Robot.} Actuators are essentially the robot's muscles, translating the information received from sensors into physical movement. In addition to hydraulic, thermal, and magnetic actuators, here are other most common types of actuators utilized in home robots:

\begin{itemize}
\item Electrical Actuators: These are the most prevalent type, utilizing electric motors to power various robot functions, from driving wheels in robotic vacuums to the complex arm movements of a robotic chef.
\item Pneumatic Actuators: Powered by compressed air, these actuators offer high power and speed, often used in applications requiring strong bursts of force, such as some lawnmowers.
\end{itemize}

\textbf{Sensory Fusion: The Power of Teamwork.} It's important to note that sensors and actuators rarely operate in isolation. A critical concept in robotics is sensory fusion, where data from multiple sensors is combined and processed to create a more comprehensive understanding of the environment. This allows robots to make more informed decisions and react to complex situations effectively. 



\section{Outlook of AI Robots at Home}
In our vision of future AI robotics, a diverse array of robots will seamlessly integrate into households, adopting either humanoid or specialized forms tailored to specific tasks. These robots will collectively possess a comprehensive understanding of their environment, facilitated by shared 3D scans of the home, environmental perception capabilities, and synchronized actuations. Moreover, these robots will be attuned to the profiles of household members, encompassing their habits, hobbies, and preferences, facilitating personalized interactions.

Communication with humans will primarily rely on advanced verbal interfaces powered by Large Language Models (LLMs), fostering intuitive and natural exchanges. Additionally, home robots will seamlessly interface with nearby edge devices, facilitating access to fog clusters or cloud-based resources, as depicted in Figure \ref{fig:framework}. Continuously updated through regular model inference updates, these robots will stay current with the latest developments, ensuring optimal performance and adaptability.



\textbf{Seamless Human-Computer Interaction (HCI):}
Interaction with home robots will become more natural and intuitive, with LLMs playing a pivotal role.  Imagine conversing with your robot using natural language, and the LLM behind the scenes interprets your intent and furnishes timely information or accomplishes tasks. Additionally, robots can even learn to interpret facial expressions and body language with the assistance of Large AI Models (LAM), fostering a more natural and engaging user experience. 

\textbf{Powerful AI Reasoning}:
AI reasoning capabilities will allow robots to make informed decisions and solve problems independently.  This could involve robots planning and executing complex tasks, adapting to unforeseen circumstances, and even learning from experience to continuously improve their performance. Envision a robot chef that can not only follow a recipe but also improvise based on missing ingredients or your personal preferences.

\textbf{Reinforcement Learning from Human Feedback (RLHF)}:
A fascinating area of research is RLHF, where robots learn through trial and error, guided by human feedback. This allows robots to personalize their behavior to each user's preferences and continuously improve their performance over time.  For example, a robotic vacuum cleaner might learn the most efficient cleaning patterns for your specific home layout based on your feedback.

\section{Challenges}
The path towards seamless AI robot integration within our homes is paved with both exciting possibilities and significant challenges. Here, we explore some key hurdles that need to be addressed:

\textbf{Communication}
For robots to effectively communicate and receive instructions, reliable and robust communication channels are essential.  However, limitations in home networks can lead to delays, dropped connections, and hindered performance. Optimizing data transfer protocols and potentially leveraging advancements in 5G/6G technology can address these bandwidth and latency bottlenecks, ensuring smooth and high QoE communication between robots and their distributed computing systems.

\textbf{Computation Resources}
The complex algorithms powering AI robots require substantial computational resources.  One approach is edge computing, where processing power resides directly on the robot itself. While this offers advantages in terms of privacy and responsiveness,  limited onboard processing power can restrict functionalities. Alternatively, cloud computing allows robots to tap into vast processing power in the cloud. This enables more complex tasks but raises concerns about latency and potential cybersecurity risks associated with data transmission.  The key lies in finding the optimal balance between edge and cloud computing, ensuring efficient processing without compromising user privacy or robot performance.

\textbf{Robot Ethics, Safety, and Privacy}
As robots become more sophisticated, ethical considerations come to the forefront. Questions surrounding safety, privacy, data security, and potential bias in AI algorithms need to be addressed. Additionally, the concept of ``red teaming" –  testing robots for potential vulnerabilities and unintended consequences – is crucial to ensure their safe and responsible operation within our homes.


\section{Final Words from an Economy Perspective}
The economic potential of home robots is tremendous. Recent surveys show most Americans spend at least two hours per week on household chores such as laundry and cleaning, even with the help of ubiquitous appliances. That is one year in one’s lifetime. Mundane household chores are “work” as two out of three surveyed say they don’t enjoy doing them. That work is unpaid, thus not counted in official GDP. The boost to the economy should be around \$500 billion in America. In addition, Americans spend nearly half an hour per day on personal care. So the economic impact of home robots could be roughly 2 percent of GDP, equal to the contribution of the automotive industry. We can do similar estimates for other countries.


\bibliographystyle{IEEEtran}
\bibliography{ref.bib}

\begin{IEEEbiography}{Haiwei Dong} is a Principal Researcher with Huawei Technologies Canada and an Adjunct Professor at the University of Ottawa. His research interests include artificial intelligence, multimedia, Metaverse, and robotics. Dong received his Ph.D. degree from Kobe University, Kobe, Japan. He is a senior member of the IEEE. Contact him at haiwei.dong@ieee.org.
\end{IEEEbiography}

\begin{IEEEbiography}{Yang Liu} is a Staff Engineer with Shopify and a PhD Candidate at the University of Ottawa. His research interests include artificial intelligence and multimedia. Yang received his Master degree from the University of Ottawa. He is a member of the IEEE. Contact him at yliu344@uottawa.ca.
\end{IEEEbiography}

\begin{IEEEbiography}{Ted Chu} is the Chief Economic Advisor with Millennium Institute. He was a Chief Economist at Saudi Public Investment Fund, World Bank, International Finance Corporation, ADIA, and GM. Contact him at  tc@Millennium-Institute.org.
\end{IEEEbiography}

\begin{IEEEbiography}{Abdulmotaleb El Saddik} is a Distinguished Professor at the University of Ottawa and MBZUAI. He is a Fellow of Royal Society of Canada, a Fellow of IEEE, an ACM Distinguished Scientist, and a Fellow of the Engineering Institute of Canada and the Canadian Academy of Engineers. Contact him at elsaddik@uottawa.ca.
\end{IEEEbiography}





\end{document}